\documentclass{article}

\usepackage{PRIMEarxiv}

\usepackage[utf8]{inputenc} 

\usepackage{lipsum}
\usepackage{fancyhdr}       
\usepackage{graphicx}       
\graphicspath{{media/}}     
\usepackage[utf8]{inputenc} 
\usepackage[T1]{fontenc}    
\usepackage[colorlinks]{hyperref}       
\usepackage{url}            
\usepackage{booktabs}       
\usepackage{bbding}
\usepackage{amsfonts}       
\usepackage{nicefrac}       
\usepackage{microtype}      
\usepackage{xcolor, soul}         
\usepackage{array,multirow,graphicx}
\usepackage{glossaries}
\usepackage{wrapfig}
\usepackage{import}
\usepackage{standalone}

\newacronym{LBP}{LBP}{local binary pattern}
\newacronym{FER}{FER}{Facial Expression Recognition}
\newacronym{CNN}{CNN}{Convolution Neural Network}
\newacronym{ViT}{ViT}{Vision Transformer}
\newacronym{NLP}{NLP}{Natural Language Processing}
\newacronym{HOG}{HOG}{Histogram of Oriented Gradients}
\newacronym{OF}{OF}{Optical Flow}
\newacronym{SE}{SE}{Squeeze and Excitation}
\newacronym{MLP}{MLP}{Multiple Layer Perceptron}
\newacronym{LPQ}{LPQ}{Local Phase Quantization}
\newacronym{PHOG}{PHOG}{Pyramid of Histogram of Oriented Gradients}
\newacronym{PCA}{PCA}{ Principal Component Analysis}
\newacronym{CCA}{CCA}{Canonical Correlation Analysis}
\newacronym{SVM}{SVM}{Support Vector Machine}
\newacronym{SPC}{SPC}{Symmetric Positive Definite}
\newacronym{MDS}{MDS}{ Multi Dimensional Scaling}
\newacronym{MHA}{MHA}{Multi-Head Attention}
\newacronym{CK}{CK}{Cohne-Kanade}
\newacronym{JAFFE}{JAFFE}{Japanese Female Facial Expression}
\newacronym{RAF-DB}{RAF-DB}{Real-world Affective Faces Database}
\newacronym{SFEW}{SFEW}{Static Facial Expression in the Wild}
\newacronym{CAM}{CAM}{Class Activation Mapping}

\title{Learning Vision Transformer with Squeeze and Excitation for Facial Expression Recognition}

\author{
  Mouath Aouayeb \\
  Univ.   Rennes,
  INSA   Rennes, CentraleSupélec,\\  CNRS,   IETR   -   UMR   6164, \\  Rennes,   France\\
  \texttt{aouayeb.mouath@insa-rennes.fr} \\
   \And
  Wassim Hamidouche \\
  Univ.   Rennes,
  INSA   Rennes,  \\ CNRS,   IETR   -   UMR   6164, \\  Rennes,   France\\ 
  \texttt{wassim.hamidouche@insa-rennes.fr} \\
  \And
  Catherine Soladie \\
  Univ.   Rennes,
  CentraleSupélec,  \\ CNRS,   IETR   -   UMR   6164, \\ Rennes,   France\\
  \texttt{catherine.soladie@centralesupelec.fr} \\
  \And
  Kidiyo Kpalma \\
  Univ.   Rennes,
  INSA   Rennes,  \\ CNRS,   IETR   -   UMR   6164, \\  Rennes,   France\\
  \texttt{kidiyo.kpalma@insa-rennes.fr} \\
  \And
  Renaud Seguier \\
  Univ.   Rennes,
  CentraleSupélec,  \\ CNRS,   IETR   -   UMR   6164, \\ Rennes,   France \\
  \texttt{renaud.seguier@centralesupelec.fr} \\
}

\begin{document}
\maketitle

\begin{abstract}
As various databases of facial expressions have been made accessible over the last few decades, the \gls{FER} task has gotten a lot of interest. The multiple sources of the available databases raised several challenges for facial recognition task. These challenges are usually addressed by \gls{CNN} architectures. Different from \gls{CNN} models, a Transformer model based on attention mechanism has been presented recently to address vision tasks. One of the major issue with Transformers is the need of a large data for training, while most \gls{FER} databases are limited compared to other vision applications. Therefore, we propose in this paper to learn a vision Transformer jointly with a \gls{SE} block for \gls{FER} task. The proposed method is evaluated on different publicly available \gls{FER} databases including CK+, JAFFE, RAF-DB and SFEW.  Experiments demonstrate that our model outperforms state-of-the-art methods on CK+ and SFEW and achieves competitive results on JAFFE and  RAF-DB.
\end{abstract}


\section{Introduction}
\label{sec:intro}
Year after year, human life is increasingly intertwined with Artificial Intelligence (AI)-based systems. As a result, there is a growing attention in technologies that can understand and interact with humans, or that can provide improved contact between humans.
To that end, more researchers are involved in developing automated \gls{FER} methods that can be summarised in three categories including Handcrafted, Deep Learning and Hybrid. Main handcrafted solutions~\cite{506414,7026204,9378702} are based on techniques like \gls{LBP}, \gls{HOG} and \gls{OF}. They present good results on lab-made databases (CK+~\cite{5543262} and JAFFE~\cite{670949}), in contrast, they perform modestly on wild databases (SFEW~\cite{6130508} and RAF-DB~\cite{li2017reliable}). Some researchers~\cite{9191181, NEURIPS2020_a51fb975,Farzaneh_2021_WACV} have taken advantage of advancements in deep learning techniques, especially in \gls{CNN} architectures, to outperform previous hand-crafted solutions. Others~\cite{article1, 9084763} propose solutions that mix the handcrafted techniques with deep learning techniques to address specific challenges in \gls{FER}.

Impressive results~\cite{NIPS2017_3f5ee243,devlin-etal-2019-bert, liu2019roberta} from Transformer models on \gls{NLP} tasks have motivated vision community to study the application of Transformers to computer vision problems. The idea is to represent an image as a sequence of patches in analogy of a sequence of words in a sentence in \gls{NLP} domain. Transformers are made to learn parallel relation between sequence inputs through an attention mechanism which makes them theoretically suitable for both tasks \gls{NLP} and image processing.
The Transformer was firstly introduced by Vaswani {\it et al.}~\cite{NIPS2017_3f5ee243} as a machine translation model, and then multiple variants~\cite{NIPS2017_3f5ee243,devlin-etal-2019-bert, liu2019roberta} were proposed to increase the model accuracy and overcome various \gls{NLP} challenges. 
Recently, a \gls{ViT} is presented for different computer vision tasks from image classification~\cite{dosovitskiy2020}, object detection~\cite{carion2020end} to image data generation~\cite{jiang2021transgan}. The Transformer proves its capability and overcomes state-of-the-art performance in different \gls{NLP} applications as well as in vision applications.  
However, these attention-based architectures are computationally more demanding than \glspl{CNN} and training data hunger.

In this paper, we propose to alleviate the problem, that \gls{ViT} has, caused by the lack of training data for \gls{FER} with a block of \gls{SE}. We also provide an internal representations analysis of the \gls{ViT} on facial expressions. 
The contribution of this paper can be summarized in four-folds:
\begin{itemize}
    \item Introduction of a \gls{SE} block to optimize the learning of the \gls{ViT}.
    \item Fine-tuning of the \gls{ViT} on FER-2013~\cite{carrier2013fer} database for \gls{FER} task.
    \item Test of the model on four different databases (CK+~\cite{5543262}, JAFFE~\cite{670949}, RAF-DB~\cite{li2017reliable}, SFEW~\cite{6130508}).
    \item Analysis of the attention mechanism of the \gls{ViT} and the effect of the \gls{SE} block. 
\end{itemize}

The remaining of this paper is organized as follows. Section~\ref{sec:Related_works} reviews the related work. Section \ref{sec:Proposed_method} firstly gives an overview of the proposed method and then describes the details of the \gls{ViT} and the \gls{SE} block. Section \ref{sec:Exp_comparison} presents the experimental results. Finally, Section \ref{sec:Conclusion} concludes the paper.

\section{Related Works}
\label{sec:Related_works}
In this section, we briefly review some related works on \gls{ViT} and facial expression recognition solutions.
\subsection{Vision Transformer (ViT)}
The \gls{ViT} is first proposed by Dosovitskiy {\it et al.}~\cite{dosovitskiy2020} for image classification. The main part of the model is the encoder part of the Transformer as first introduced for machine translation by Vaswani {\it et al.}~\cite{NIPS2017_3f5ee243}. To transform the images into a sequence of patches they use a linear projection, and for the classification, they use only the token class vector. The model achieves state-of-the-art performance on ImageNet~\cite{5206848} classification using fine-tuning on JFT-300M~\cite{8237359}. From that and the fact that this model contains much more parameters (about 100M) than \glspl{CNN}, we can say that \gls{ViT} are data-hungry models. To address this  heavily relying on large-scale databases, Touvron {\it et al.}~\cite{touvron2020deit} proposed DEIT model. It's a \gls{ViT} with two classification tokens. The first one is fed to an \gls{MLP} head for the classification and the other one is used on the distillation process with a \gls{CNN} teacher model pretrained on ImageNet~\cite{5206848}. The DEIT was only trained on ImageNet and outperforms both the \gls{ViT} model and the teacher model. Yuan {\it et al.}~\cite{yuan2021tokens} overcome the same limitation of \gls{ViT} using novel tokenization process. The proposed T2T-ViT~\cite{yuan2021tokens} model has two modules: 1) the T2T tokenization module that consists in two steps: re-structurization and soft split, to model the local information and reduce the length of tokens progressively, and 2) the Transformer encoder module. It achieves state-of-the-art performance on ImageNet~\cite{5206848} classification without a pretraining on JFT-300M~\cite{8237359}.

\subsection{Facial Expression Recognition}
The \gls{FER} task has progressed from handcrafted \cite{506414,7026204,9378702} solutions to deep learning \cite{9191181,Otberdout2018DeepCD,Farzaneh_2021_WACV,Wang2020RegionAN} and Hybrid \cite{article1,9084763,Ma2021RobustFE} solutions. 
In 2014, Turan {\it et al.} \cite{7026204} proposed a region-based handcrafted system for \gls{FER}. They extracted features from the eye and mouth regions using \gls{LPQ} and \gls{PHOG}. A \gls{PCA} is used as a tool for features selection. They fused the two groups of features with a \gls{CCA} and finally, a \gls{SVM} is applied as a classifier. More recent work~\cite{9378702}, proposed an automatic \gls{FER} system based on \gls{LBP} and \gls{HOG} as features extractor. A local linear embedding technique is used to reduce features dimensionality and a \gls{SVM} for the classification part. They reached state-of-the-art performance for handcrafted solutions on JAFFE~\cite{670949}, KDEF~\cite{kdef51} and RafD~\cite{rafd52}.
Recently, more challenging and rich data have been made publicly available and with the progress of deep learning architectures, many deep learning solutions based on \gls{CNN} models are revealed. Otberdout {\it et al.} \cite{Otberdout2018DeepCD} proposed to use \gls{SPC} to replace the fully connected layer in \gls{CNN} architecture for facial expression classification. Wang {\it et al.} \cite{Wang2020RegionAN} proposed a region-based solution with a \gls{CNN} model with two blocks of attention. They perform different crop of the same image and apply a \gls{CNN} on each patch. A self-attention module is then applied followed by a relation attention module. On the self-attention block, they use a loss function in  a way that one of the cropped image may have a weight larger than the weight given to the input image. More recently, Farzaneh {\it et al.} \cite{Farzaneh_2021_WACV} have integrated an attention block to estimate the weights of features with a sparse center loss to achieve intra-class compactness and inter-class separation. Deep learning based solutions have widely outperformed handcrafted solutions especially on wild databases like RAF-DB~\cite{li2017reliable}, SFEW\cite{6130508}, AffectNet~\cite{8013713} and others. 

Other researchers have though about combining deep learning techniques with handcrafted techniques into a hybrid system. Levi {\it et al.} \cite{article1} proposed to apply \gls{CNN} on the image, its \gls{LBP} and the mapped \gls{LBP} to a 3D space using \gls{MDS}. Xu {\it et al.}~\cite{9084763} proposed to fuse \gls{CNN} features with \gls{LBP} features and they used \gls{PCA} as features selector. Newly, many Transformer models have been introduced for different  computer vision tasks and in that context Ma {\it et al.} \cite{Ma2021RobustFE} proposed a convolutional vision Transformer. They extract features from the input image as well as form its \gls{LBP} using a ResNet18. Then, they fuse the extracted features with an attentional selective fusion module and fed the output to a Transformer encoder with a \gls{MLP} head to perform the classification. To our knowledge, \cite{Ma2021RobustFE} is considered as the first solution based on Transformer architecture for \gls{FER}. However, our proposed solution differs in applying the Transformer encoder directly on the image and not on the extracted features which may reduce the complexity of the proposed system and aid to study and analyse the application of \gls{ViT} on \gls{FER} problem as one of the interesting vision tasks. 

Table~\ref{tab:SOTA} (presented in the Supplementary Material) summarizes some state-of-the-art approaches with details on the used architecture and databases. We can notice that different databases are used to address different issues and challenges. From these databases we selected 4 of them to study our proposed solution and compare it with state-of-the-art works. The selected databases are described in the experiments and comparison Section~\ref{sec:Exp_comparison}. In the next section we will describe our proposed solution. 

\section{Proposed Method}
\label{sec:Proposed_method}
In this section, we introduce the proposed solution in three separate paragraphs: an overview, then some details of the \gls{ViT} architecture and the attention mechanism, and finally the \gls{SE} block. 
\subsection{Architecture overview}
The proposed  solution contains two main parts, a vision Transformer to extract local attention features and a \gls{SE} block to extract global relation from the extracted features which may optimize the learning process on small facial expressions databases.

\begin{figure}[!ht]
    \centering
    \includegraphics[width=\textwidth]{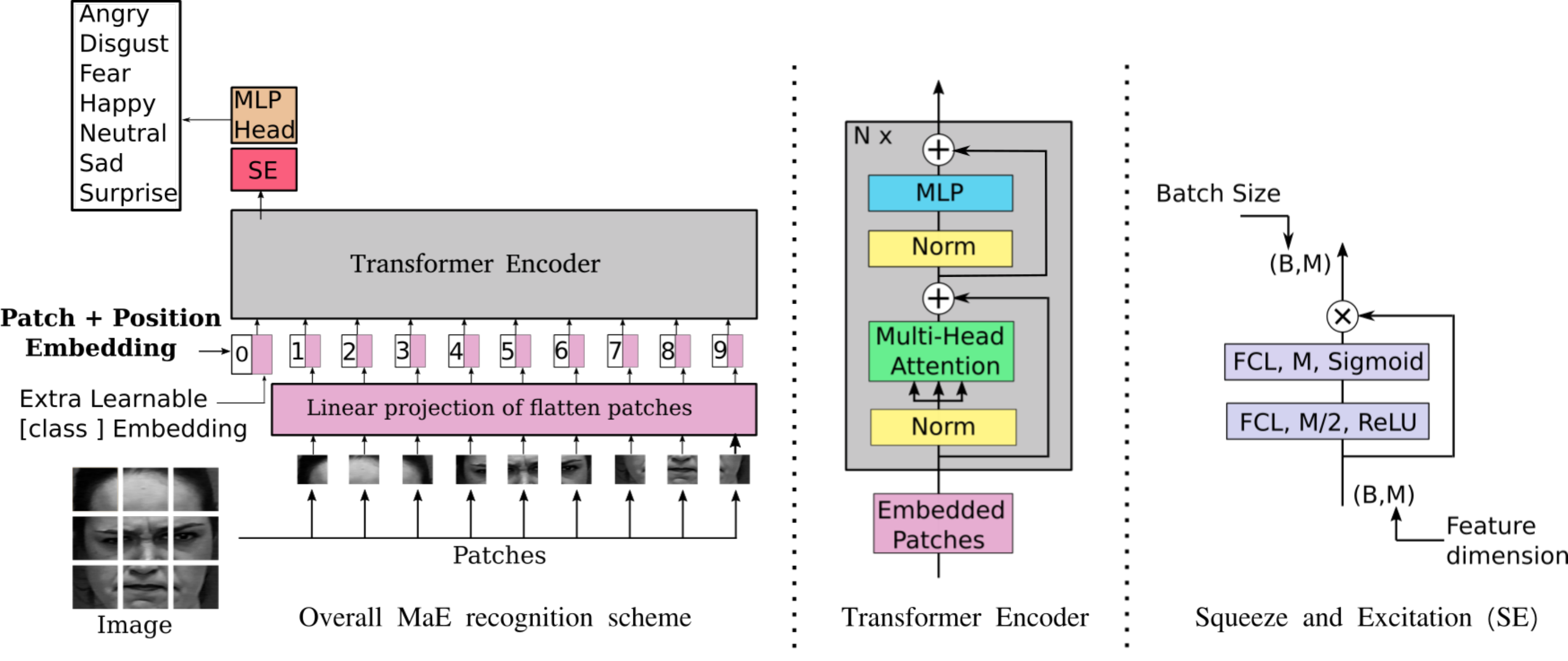}
    \caption{Overview of the proposed solution. The used \gls{ViT} is the base version with 14 layers of Transformer encoder and patch dimension of $(16 \times 16)$. The \gls{ViT} is already trained on JFT-300M~\cite{8237359} database and fine-tuned to ImageNet-1K~\cite{5206848} database. 
    }
    \label{fig:overview}
\end{figure}
\subsection{Vision Transformer}
The vision Transformer consists of two steps: the tokenization and the Transformer encoder. In the tokenization step, the image is cropped onto $L$ equal $(h \times h)$ dimension patches and then flattened to a vector. An extra learnable vector is added as a token for classification called "cls\_tkn". Each vector is marked with a position value. To summarize, the input of the Transformer encoder is $L+1$ vectors of length $h^2+1$. 

As shown in Figure \ref{fig:overview}, the Transformer encoder is a sequence of $N$ blocks of the attention module. The main part of the attention block is the \gls{MHA}. The \gls{MHA} is build with $z$ heads of self-Attention, also called intra-attention. According to \cite{NIPS2017_3f5ee243}, the idea of the self-attention is to relate different positions of a single sequence in order to compute a representation of the sequence.  
For a given sequence, 3 layers are used: Q-layer, K-layer and V-layer and the self-attention function will be a mapping of a query (Q or Q-layer) and a set of key-value (K or K-layer; V or V-layer) pairs to an output. The self-attention function is summarized by Equation~\eqref{eq:attention}:
\begin{equation}
    \begin{split}
    Attention(Q,K,V) &= softmax(\frac{QK^{T}}{\sqrt{d_{k}}})V.
    \end{split}
    \label{eq:attention}
\end{equation}
And so the \gls{MHA} Equation \eqref{eq:mha} will be: 
\begin{equation}
    \begin{split}
    MHA(Q,K,V) &= Concat(head_{0},...,head_{z})W^{O}, \\
     head_{i} &= Attention(QW_{i}^{Q},KW_{i}^{K},VW_{i}^{V}).
    \end{split}
    \label{eq:mha}
\end{equation}
where the projections $W^{O}, W_{i}^{Q}, W_{i}^{K}$ and $W_{i}^{V}$ are parameters' matrices.

\subsection{Squeeze and Excitation (SE)}
The Squeeze and Excitation block, shown on the right of the Figure \ref{fig:overview}, is also an attention mechanism. 
It contains widely fewer parameters than self-attention block as shown by Equation~\eqref{eq:se} where two fully connected layers are used with only one operation of pointwise multiplication. It is firstly introduced in \cite{8578843} to optimize \gls{CNN} architecture as a channel-wise attention module, concretely we use only the excitation part since the squeeze part is a pooling layer build to reduce the dimension of the 2d-CNN layers.
\begin{equation}
    \begin{split}
        SE(cls\_tkn) &= cls\_tkn \odot  Excitaion(cls\_tkn), \\ 
        Excitaion(cls\_tkn) &= Sigmoid(FCL_{\gamma}(ReLU(FCL_{\gamma/4}(cls\_tkn)))).
    \end{split}
    \label{eq:se}
\end{equation}
where $FCL_{\gamma}$ and $FCL_{\gamma/4}$ are fully connected layers with respectively $\gamma$ neurons and $\gamma/4$ neurons, $\gamma$ is the length of the cls\_tkn which is the classification token vector and $\odot $ is a pointwise multiplication. 
The idea of using \gls{SE} in our architecture is to optimize the learning of the \gls{ViT} by learning more global attention relations between extracted local attention features. Thus, the \gls{SE} is introduced on top of the Transformer encoder more precisely on the classification token vector. Different from the self-attention block where it is used inside the Transformer encoder to encode the input sequence and extract features through cls\_tkn, the \gls{SE} is applied to recalibrate the feature responses by explicitly modelling inter-dependencies among cls\_tkn channels.   
\section{Experiments and Comparison}
\label{sec:Exp_comparison}
In this section, we first describe the used databases, and then provide an ablation study for different contributions with other details on the proposed solution and an analysis of additional visualisation for in-depth understanding of the \gls{ViT} applied on \gls{FER} task. Finally, we present a comparison with state-of-the-art works.
\subsection{FER Databases}
\textbf{CK+}~\cite{5543262} : published on 2010, and it is an extended version of \gls{CK} database. It contains 593 sequences taken in lab environment with two data formats $(640 \times 490)$ and $(640 \times 480)$. It encompasses the 7 basic expressions which are : Angry, Disgust, Fear, Happy, Neutral, sad and Surprise, plus the Contempt expression. In our case, we only worked on the 7 basic expressions to have a fair study with other databases and with most state-of-the-art solutions.  \\
\textbf{JAFFE}~\cite{670949}: The \gls{JAFFE} database is a 213 gray scale images of acted Japanese female facial expressions. All the images are resized onto $(256 \times 256)$. It contains the 7 basic expressions. \\
\textbf{FER-2013}~\cite{carrier2013fer}: The FER-2013 database, or sometimes referred as FERPlus, is almost  35k facial expressions database on 7 basic expressions. Published in 2013 in a challenge on Kaggle plate-form\footnote{https://www.kaggle.com/msambare/FER-2013}. The images are collected from the web converted to gray scale model and resized to $(48 \times 48)$. Theoretically, this database could suffer from  mislabeling since a $68\% \pm 5\%$ human accuracy is reported. However, since it is a large spontaneous databases of facial expressions we used it as a pre-training data for our model.  \\
\textbf{SFEW}~\cite{6130508}: The \gls{SFEW} is a very challenging databases with images captured from different movies. It contains 1,766 RGB images with size of $(720 \times 576)$. It is also labeled with the 7 basic expressions.  \\
\textbf{RAF-DB}~\cite{li2017reliable}: The \gls{RAF-DB} is a recent database with nearly 30K of mixed RGB and gray scale images collected from different internet websites. This database contains two separate sub-data: one with 7 basic expressions and the other with  12 compound facial expressions. In the experiments, we used the 7 basic expressions version.

Table~\ref{tab:databases} (presented in the Supplementary Material) summarizes previous presented databases with reference to the year and the publication conference and some other details. For \gls{FER} task there are other publicly available databases that address different issues, but we restrained our choices on these databases because they are in the center of interest of major state-of-the-art solutions. 
\subsection{Architecture and training parameters}
In all experiments, we use a pretrained ViT-B16-224 (weights\footnote{\url{https://github.com/rwightman/pytorch-image-models/blob/master/timm/models/vision_transformer.py}}), the base version of the \gls{ViT} with $(16 \times 16)$ patch size and $(224 \times 224)$ input image size. Since \gls{ViT} training needs large data to reach good performance we used the following list of data augmentation:  Random Horizontal flip, Random GrayScale conversion, different values of brightness, contrast and saturation. All images are converted to 3 channels, resized to $(224 \times 224)$ and normalized. The regularisation methods we used in this work are Cutout~\cite{devries2017cutout} and Mixup~\cite{zhang2018mixup}. The training is performed with categorical cross entropy as a loss function and AdamW~\cite{Loshchilov2019DecoupledWD} as an optimizer. 
The learning rate is fixed to $1.6 \times 10^{-4}$ with a batch size of 16. When training on FER-2013 database, the number of epochs is fixed to 8 and for the rest of databases it is fixed to 10. The training process is carried-out on a Tesla K80 TPU with 8 cores using Pytorch1.7. 

\subsection{Ablation Study }
In the ablation study, we assess the performance of the \gls{ViT} architecture, the added \gls{SE} block and the use of FER-2013~\cite{carrier2013fer} as a pre-training data. Table~\ref{tab:Contributions} shows the result of different experiments on CK+, JAFFE, RAF-DB and SFEW.
\begin{table}[!ht]
  \caption{Ablation Study}
  \label{tab:Contributions}
  \centering
  \begin{tabular}{l|c|cccc}
    \toprule
    Model & Pre-train   & CK+~\cite{5543262} & JAFFE~\cite{670949}  & RAF-DB~\cite{li2017reliable} &SFEW~\cite{6130508}   \\
    \midrule
    \midrule
    ViT & \XSolidBrush*& 0.9857 & 0.8823 &  0.8595& 0.3828\\
    ViT + SE & \XSolidBrush*  & 0.9949 & 0.9061 & 0.8618& 0.4084\\ 
    ViT & FER-2013\cite{carrier2013fer}* & 0.9817 & \textbf{0.9483} & 0.8703 & 0.5035\\ 
    ViT + SE  & FER-2013\cite{carrier2013fer}*  & \textbf{0.9980} & 0.9292 & \textbf{0.8722} & \textbf{0.5429}\\
    \bottomrule
    
  \end{tabular}
  
  * The used ViT model is already trained on ImageNet~\cite{5206848}.
\end{table}
From first line, we can notice that \gls{ViT} can reach state-of-the-art performance on lab-made databases like CK+~\cite{5543262} and JAFFE~\cite{670949}, however on SFEW~\cite{6130508} the Transformer is less effective. 
In all cases, we can notice that there is a benefit of using \gls{SE} and the pre-training phase on FER-2013 \cite{carrier2013fer}. The two contributions may not be complementary on lab-made data (CK++~\cite{5543262} and JAFFE~\cite{670949}). For example, on CK++~\cite{5543262} we can notice that the pre-training improves the performance only when combined with the \gls{SE}. On JAFFE~\cite{670949}, the best solution is the one that relies on pre-training without the \gls{SE}. Although, on wild databases (RAF-DB~\cite{li2017reliable} and SFEW~\cite{6130508}) the added value of both contributions is more noticeable, specially on SFEW~\cite{6130508} we can obtain a 16\% gain on accuracy compared to the \gls{ViT} without a \gls{SE} neither a pre-training on FER-2013 \cite{carrier2013fer}.  
\begin{figure}[!ht]
    \centering
    \includegraphics[width=0.85\textwidth]{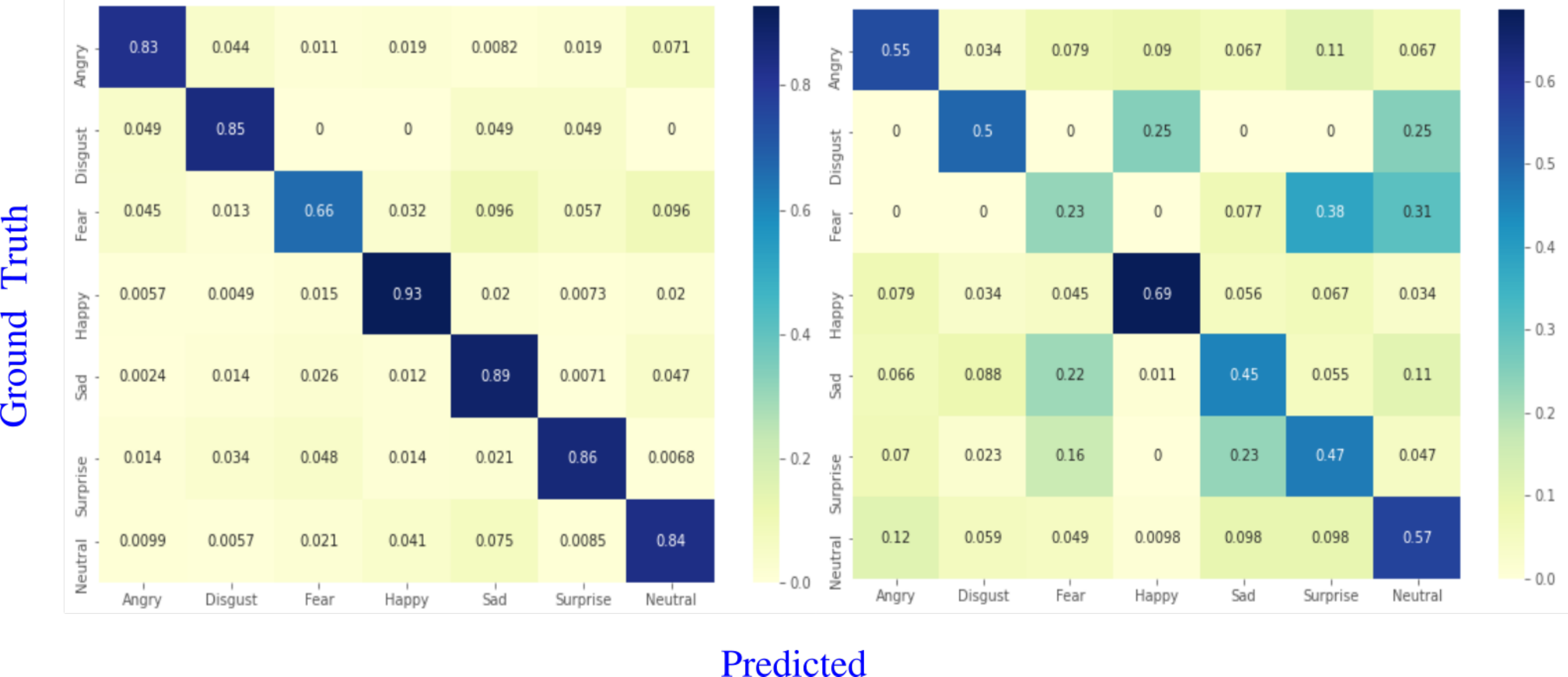}
    \caption{Confusion matrices of ViT+SE on the validation set of RAF-DB (left) and the validation set of SFEW (right).}
    \label{fig:cmx}
\end{figure}

The confusion matrices of the proposed \gls{ViT}+\gls{SE} pre-trained on FER-2013 are reported in Figure~\ref{fig:cmx}, the left plot is for the validation set of RAF-DB~\cite{li2017reliable} and the right plot is for the validation set of SFEW~\cite{6130508}. The Happy and Neutral expressions are the best recognized on the SFEW~\cite{6130508} database with respectively an accuracy of 69\% and 57\%. For RAF-DB~\cite{li2017reliable}, the Happy expression has the best accuracy with 93\% followed by the Sad expression with 89\% accuracy. On the two confusion matrices, we can notice that our model confront difficulties in recognizing the Fear expression, and that may be due to the less amount of data provided for that expression compared to the rest of expressions.
\subsection{Transformer visualisation and analysis}
In this section, we have conducted a various set of experiments in \gls{RAF-DB} database. Specially, we evaluate the classification outputs of the model through t-SNE and we provide a visual analysis of the ViT model performance with the \gls{SE} in comparison with \gls{CNN}.
\begin{figure}
    \centering
    \includegraphics[width=\textwidth]{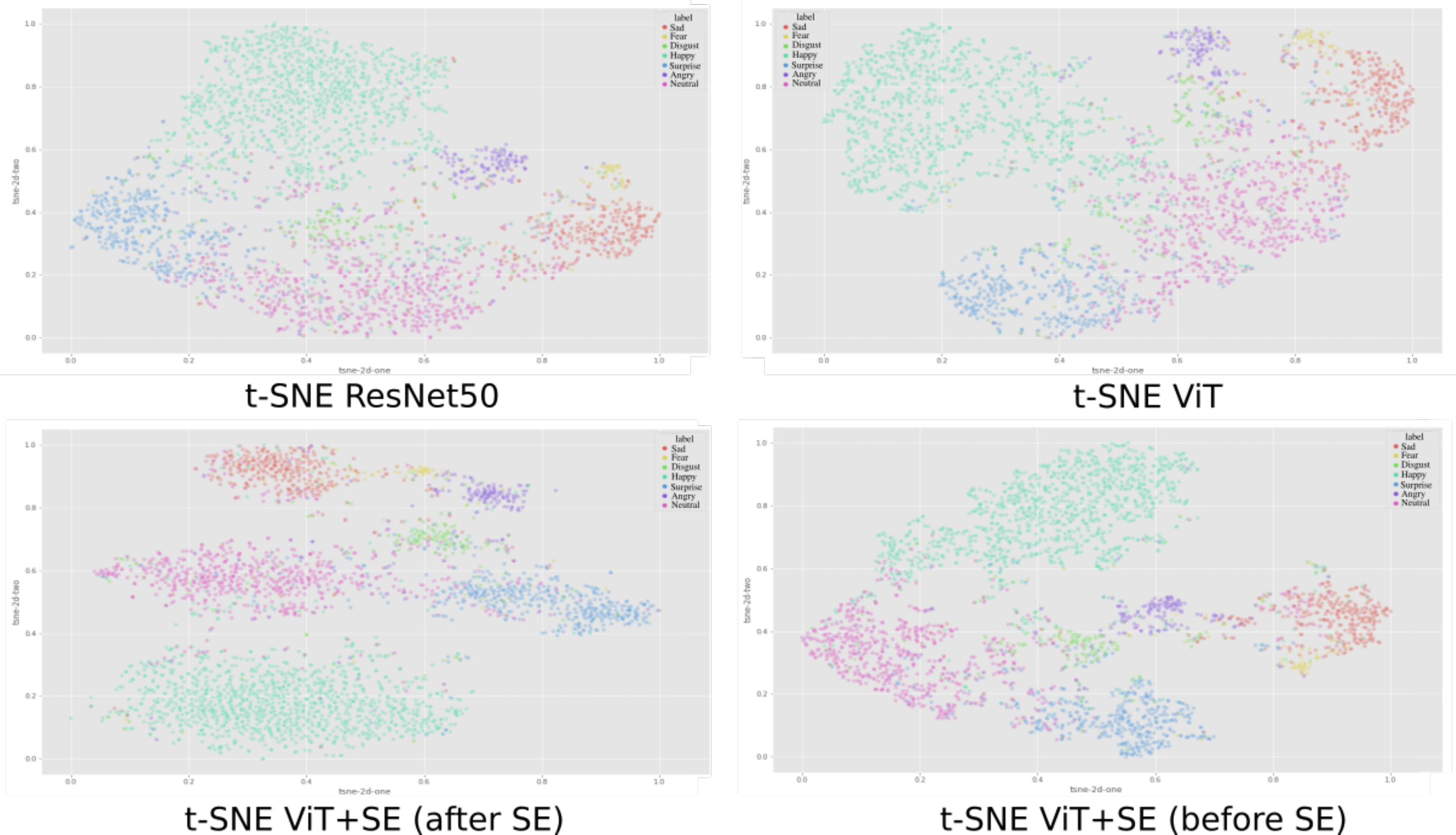}
    \caption{t-SNE plots corresponding to the 768-dimensional features from the \gls{ViT}, \gls{ViT}+\gls{SE} before and after the \gls{SE} block and the 512-dimensional features from the ResNet50. The features correspond to the RAF-DB images. The accuracy of ResNet50, ViT and ViT+SE on RAF-DB are respectively: 0.8061, 0.8595 and 0.8618. }
    \label{fig:TSNE}
     \vspace{-20pt}
\end{figure}

Figure~\ref{fig:TSNE} shows the t-SNE of the extracted features form the \gls{ViT} model without \gls{SE}, the features of the \gls{ViT} + \gls{SE} after the \gls{SE} block and before \gls{SE}, and compared with t-SNE of ResNet50~\cite{7780459} features trained also on \gls{RAF-DB}. Based on t-SNE, the \gls{ViT} architectures enable better separation of classes compared to \gls{CNN} base-line architecture (ResNet50). 
\begin{figure}
    \centering
    \includegraphics[width=0.5\textwidth]{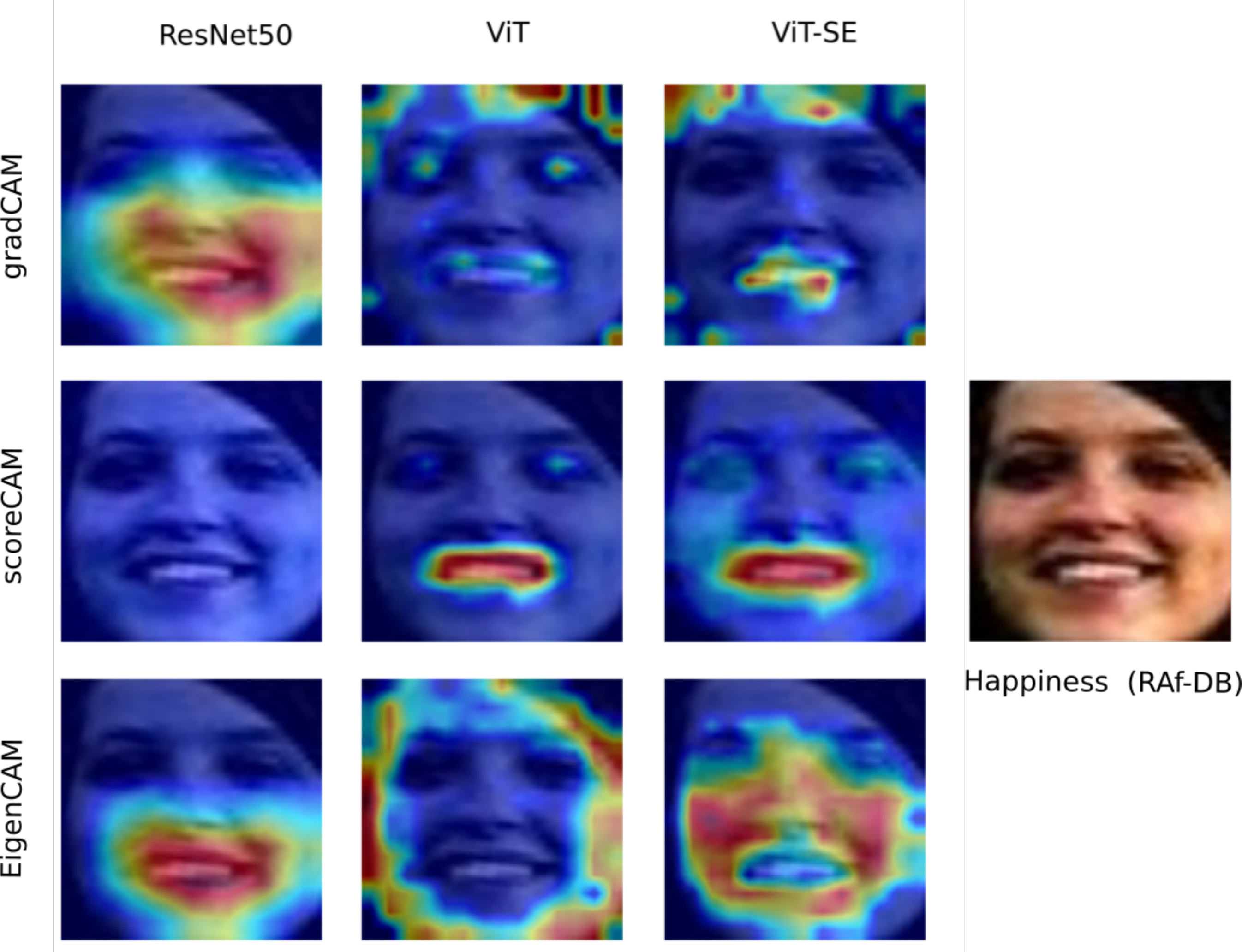}
    \caption{GRAD-CAM, Score-CAM, Eigen-CAM maps of the last layer before the classification block for the Happy expression (image from the validation set of RAF-DB~\cite{li2017reliable}).}
  \label{fig:CAM}
\end{figure}
In addition, the \gls{SE} block enhances ViT model robustness, as the intra-distances between clusters are maximized. Interestingly, the features before the \gls{SE} form a more compact clusters with inter-distance lower than the features after the \gls{SE}, which may interpret the features before \gls{SE} are more robust than those after the \gls{SE}. However, we tried to use the before \gls{SE} features directly in the classification task and no performance gain has been reported. 
Figure~\ref{fig:CAM} shows different maps of attention of the \gls{ViT}, the \gls{ViT}+\gls{SE} and the ResNet50, using Grad-Cam~\cite{8237336}, Score-Cam~\cite{9150840} and Eigen-Cam~\cite{Muhammad2020EigenCAMCA} tools. This visualisation shows that \gls{ViT} architectures succeed to focus more locally which confirm the interest of using the self-attention blocks for computer vision tasks. Once again, we can notice the gain of using the \gls{SE} block with different tools but mostly using Eigen-CAM~\cite{Muhammad2020EigenCAMCA}.

\begin{figure}[!ht]
    \centering
    \includegraphics[width=0.7\textwidth]{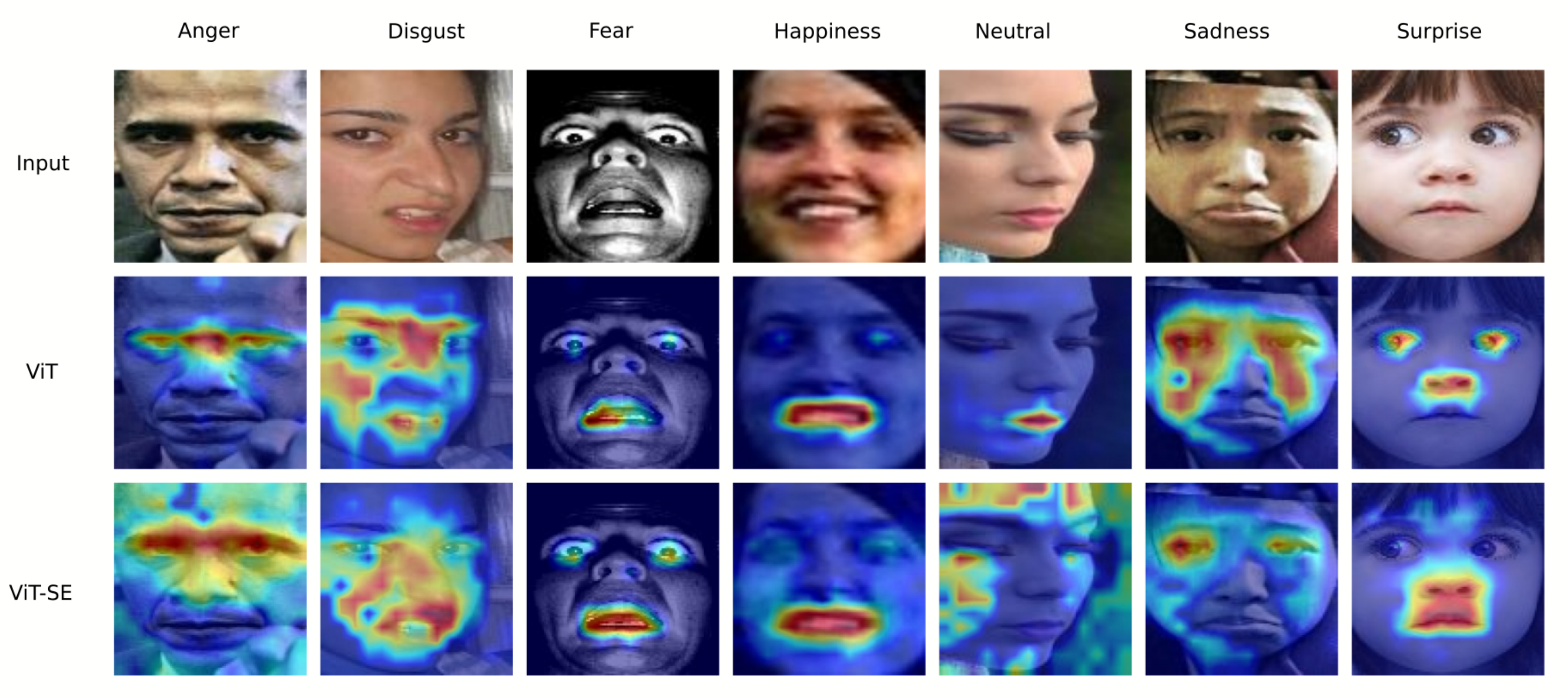}
    \caption{ Attention maps based on GRAD-CAM for different expressions (images from the validation set of RAF-DB).}
    \label{fig:Dexps}
\end{figure}

Other investigations of the \gls{ViT} architecture are presented in the Supplementary Material (Figure~\ref{fig:Dlayers}) that shows the evolution of the attention form first attention block to a deeper attention blocks and we can notice that the focus of the \gls{ViT} goes from global attention to more local attention. This particular behaviour of the \gls{ViT} on \gls{FER} task is the motivation of using \gls{SE} block on top of it to build a calibrated relation between different local focuses. In Figure~\ref{fig:Dexps} (Supplementary Material), we show the focus of the \gls{ViT} compared to the \gls{ViT} + \gls{SE} for different facial expressions and it shows how the \gls{SE} can rectify the local attention feature extracted with the \gls{ViT}, by searching for a global attention relations. 
\subsection{Comparison with state-of-the-art}
In this paper, we compare our proposed model \gls{ViT}+\gls{SE} pre-trained on FER-2013~\cite{carrier2013fer} database with state-of-the-art solution on 2 lab-made databases (CK+~\cite{5543262} and JAFFE~\cite{670949}) and 2 wild databases (RAF-DB~\cite{li2017reliable} and SFEW~\cite{6130508}).
Table~\ref{tab:comparison_ck+} shows that we have the highest accuracy on CK+~\cite{5543262} with a 99.80\% using a 10-fold cross-validation protocol. Table~\ref{tab:comparison_sfew} shows that we set the new state-of-the-art performance for single models on SFEW~\cite{6130508} with 54.29\% accuracy, however a higher accuracy (56.4\%) is reported in \cite{Wang2020RegionAN} using ensemble models. Furthermore, in Table~\ref{tab:comparison_jaffe} the proposed solution have a good 10-fold cross validation accuracy on JAFFE~\cite{670949} with 92.92\%. To our knowledge, it is the highest performance with a deep learning based solution but still less by almost 3\% than the highest obtained accuracy with newly handcrafted proposed solution~\cite{9378702}. Table~\ref{tab:comparison_rafdb} shows that our solution has a good result on RAF-DB~\cite{li2017reliable} with an accuracy of 87.22\%, to position as the third best solution among state-of-the-art on this database, less than the best record by nearly 3\%. 
\begin{table}[!ht]
    \begin{minipage}[r]{.46\linewidth}
        \centering
        \caption{Comparison on CK+~\cite{5543262} with 10-fold cross validation.}
  \label{tab:comparison_ck+}
      \begin{tabular}{r|lc}
    \toprule
    Ref. &  Model Type  & Accuracy   \\
    \midrule
    \midrule
     \cite{7026204} 2014& Handcrafted  & 0.9503\\
     \cite{NEURIPS2020_a51fb975} 2020& Deep Learning & 0.9759 \\
     \cite{Minaee2021DeepEmotionFE} 2021 & Deep Learning & 0.9800 \\
    \midrule
    ViT + SE  & Deep Learning  & \textbf{0.9980} \\
    \bottomrule
  \end{tabular}
    \end{minipage}
     \hfill%
    \begin{minipage}[l]{.47\linewidth}
        \centering
        \caption{Comparison on JAFFE~\cite{670949} with 10-fold cross validation.}
  \label{tab:comparison_jaffe}
    \begin{tabular}{r|lc}
    \toprule
    Ref. & Model Type  &  Accuracy \\
    \midrule
    \midrule
    \cite{6998925} 2015 & Handcrafted  & 0.9180 \\
     \cite{9378702} 2020 & Handcrafted & \textbf{0.9600} \\
     \cite{Minaee2021DeepEmotionFE} 2021& Deep Learning & 0.9280 \\
    \midrule
    ViT + SE  & Deep Learning & 0.9292 \\
    \bottomrule
  \end{tabular}
    \end{minipage}
\end{table}
\begin{table}[!ht]
    \begin{minipage}[r]{.46\linewidth}
        \centering
        \caption{Comparison on the validation set of RAF-DB~\cite{li2017reliable}}
        \label{tab:comparison_rafdb}
      \begin{tabular}{r|cc}
    \toprule
    Ref. & Model Type  & Accuracy  \\
    \midrule
    \midrule
      \cite{Wang2020RegionAN} 2020 & Deep Learning  & 0.8690 \\
      \cite{Ma2021RobustFE} 2021 & Hybrid  & 0.8814 \\
      \cite{Shi2021LearningTA} 2021 & Deep Learning  & \textbf{0.9055} \\
    \midrule
    ViT + SE  & Deep Learning & 0.8722\\
    \bottomrule
  \end{tabular}
    \end{minipage}
     \hfill%
    \begin{minipage}[l]{.47\linewidth}
        \centering
         \caption{Comparison on the validation set of SFEW~\cite{6130508}}
        \label{tab:comparison_sfew}
     \begin{tabular}{r|lc}
    \toprule
    Ref. & Model Type  & Accuracy  \\
    \midrule
    \midrule
    \cite{Otberdout2018DeepCD} 2018 & Deep Learning  &0.4918 \\
    \cite{Cai2018IslandLF} 2018 & Deep Learning  & 0.5252 \\
      \cite{Wang2020RegionAN} 2020 & Deep Learning  & 0.5419\\
    \midrule
    ViT + SE  & Deep Learning & \textbf{0.5429}\\
    \bottomrule
  \end{tabular}
    \end{minipage}
\end{table}
\section{Conclusion}
\label{sec:Conclusion}
In this work, we introduced the \gls{ViT}+\gls{SE}, a simple scheme that optimize the learning of the \gls{ViT} by an attention block called Squeeze and Excitation. It performs impressively well for improving the performance of \gls{ViT} in \gls{FER} task. Furthermore, it also improves the robustness of the model as shown in the t-SNE representation of the extracted features and in the attention maps. We have presented the classification performance on lab-made databases (CK+ and JAFFE) and wild databases (RAF-DB and SFEW) to evaluate the gain of the \gls{SE} block and the use of FER-2013 as a pre-training database. By comparing to different state-of-the-art solutions, we have shown that our proposed solution achieves the highest performance with a single model on CK+ and SFEW, and competitive results on JAFFE and RAF-DB. As future work, we aim to extend the \gls{ViT} architecture to address the temporal aspect for a more competitive task like micro-expressions recognition.


\nocite{*}
\bibliographystyle{unsrt} 
\bibliography{ms}

\newpage
{
\global
\begin{raggedright}
{\bfseries\sffamily{\Large Learning Vision Transformer with Squeeze and Excitation for Facial Expression Recognition (Supplementary Material)}\par}
\vskip 1.5em
\end{raggedright}
\vskip\baselineskip
  \hrule
  \vskip\baselineskip
 }



\begin{abstract}
In this supplementary material, we give further details on the conducted experiments and present a summary of the state-of-the-art solutions. In particular, we provide a visual illustrations attention maps for different expressions and at different attention layers. Besides, we support our set of experiments with confusion matrices on RAF-DB and cross database evaluation on CK+. Finally, we provide additional tables that summarize both state-of-the-art solutions and used databases.
\end{abstract}

\setcounter{section}{0}
\section{Cross-database evaluation and visual illustrations}
\begin{figure}[!ht]
    \centering
    \includegraphics[width=\textwidth]{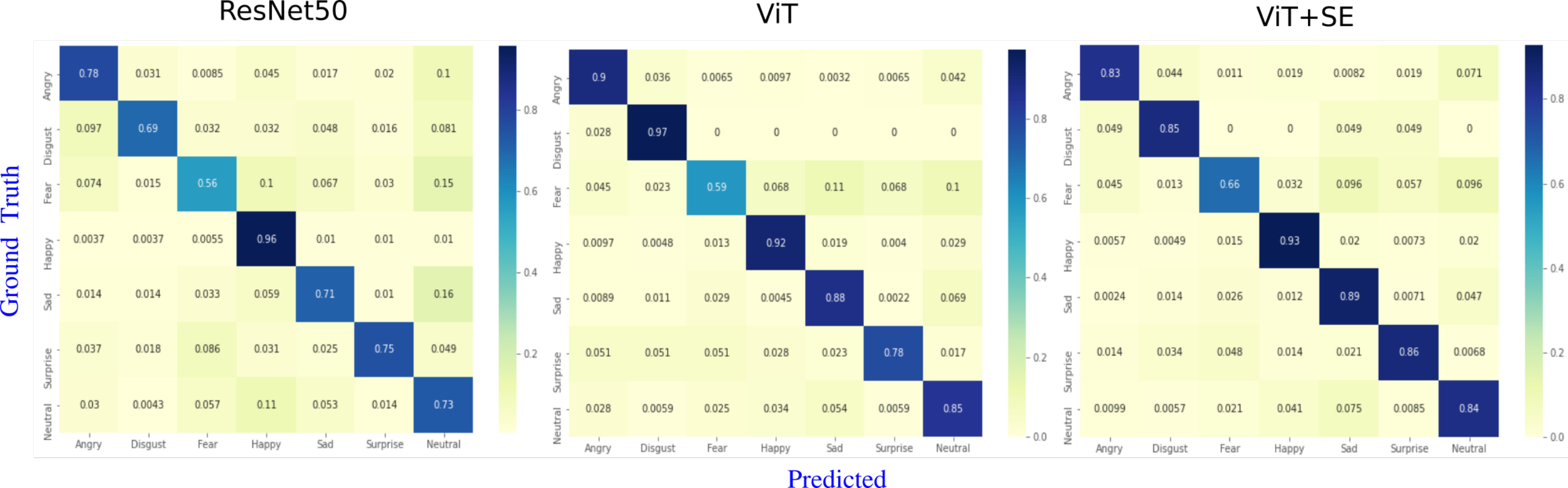}
    \caption{Confusion Matrices of RAF-DB for ResNet50 (0.8061), ViT (0.8703) and ViT+SE (0.8722).}
    \label{fig:CMs}
\end{figure}

\textbf{Cross-database evaluation:} To verify the generalisation ability of our model, we conduct a cross-database evaluation on CK+. The results are summarized in Table~\ref{tab:crossDB}. It shows that the ViT generalizes better than a baseline CNN (ResNet50), and the proposed ViT+SE model enables the best generalization from different training databases when tested on CK+. However, the generalization ability is still modest and we aim to improve it in a future work.  
\begin{table}[!ht]
\caption{Crass-database evaluation  on CK+.}
 \label{tab:crossDB}
    \centering
    \vspace{1mm}
    \begin{tabular}{l|lc|c}
    \toprule
         Model & Train & Test & Accuracy \\
         \midrule
         \multirow{4}{*}{ResNet50}  & CK+ & CK+ & 0.9488\\
          & RAf-DB & CK+ & 0.3517\\
          & SFEW & CK+ & 0.2905\\
          & FER2013 & CK+ & 0.3456\\
         \midrule
         \multirow{4}{*}{ViT} & CK+& CK+ & 0.9817\\
         & RAf-DB & CK+ & 0.5443\\
         & SFEW & CK+ & 0.3812\\
         & FER2013 & CK+ & 0.4098\\
        \midrule
        \multirow{4}{*}{ViT+SE}  & CK+ & CK+ & {\bf0.9980 }\\
         & RAf-DB & CK+ & {\bf 0.5576}\\
         & SFEW & CK+ & {\bf 0.5341 }\\
         & FER2013 & CK+ & {\bf 0.6514}\\
         \bottomrule
    \end{tabular}
\end{table}
\\
\textbf{Attention Maps:} In this work, we used Grad-Cam~\cite{8237336}, Score-Cam~\cite{9150840} and Eigen-Cam~\cite{Muhammad2020EigenCAMCA} as tools to provide visual analysis of the proposed deep learning architectures. (code available in\footnote{\url{https://github.com/jacobgil/pytorch-grad-cam}}). \\
\textbf{Grad-CAM~\cite{8237336} :} the Gradient-weighted \gls{CAM} uses the gradient of any target following to the selected layer in the model to generate a heat map that highlight the important region in the image for predicting the target. \\
\textbf{Score-CAM~\cite{9150840} :} the Score-weighted \gls{CAM} is a linear combination of weights and activation maps. The weights are obtained by passing score of each activation map forward on target class.\\
\textbf{Eigen-CAM~\cite{Muhammad2020EigenCAMCA} :} it computes the principal components of the learned features from the model layers.
\begin{figure}[!ht]
    \centering
    \includegraphics[width=0.7\textwidth]{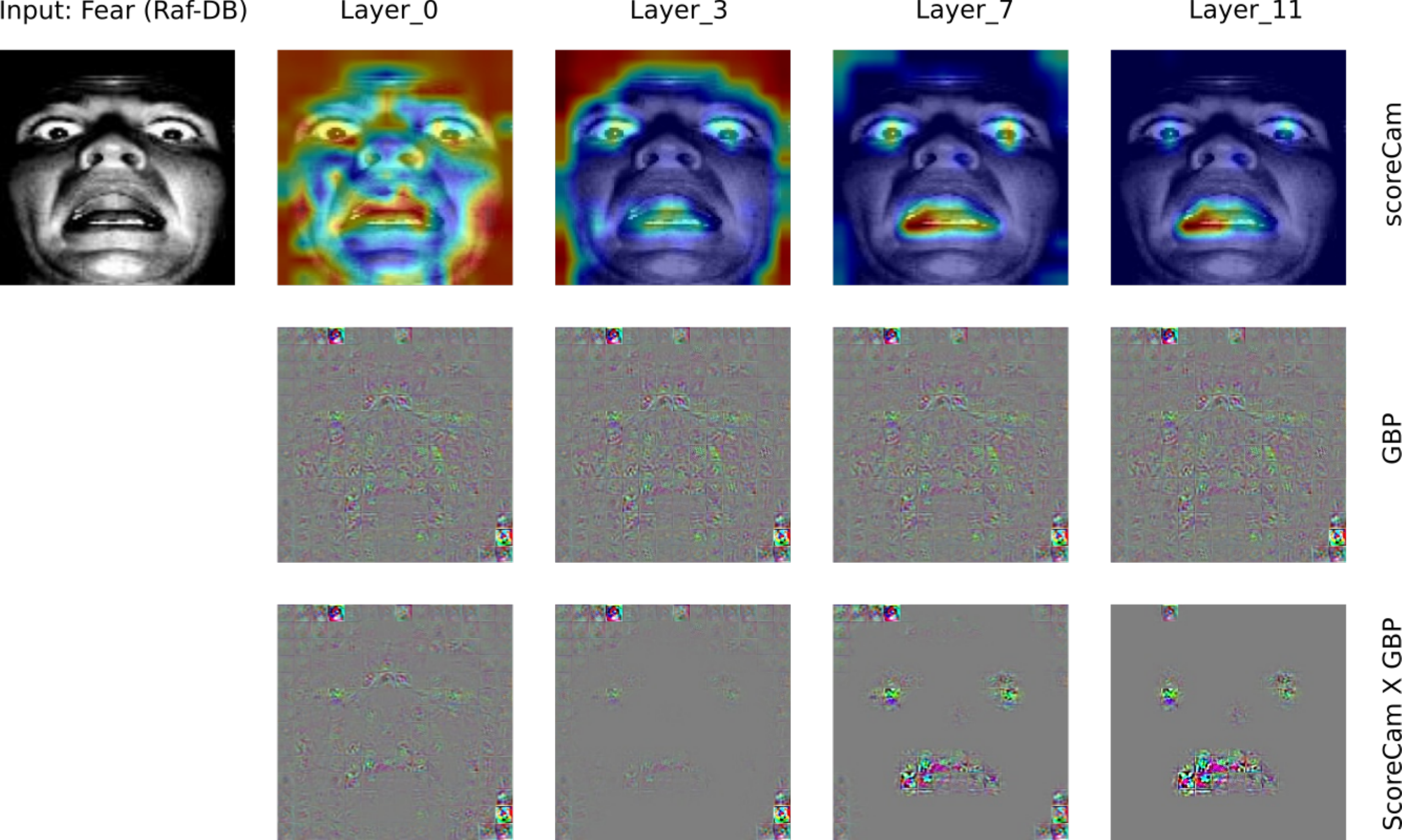}
    \caption{score-CAM maps and the guided back-propagation (GBP) at different layers of attention of the \gls{ViT} for fear expression (image from the validation set of RAF-DB).}
    \label{fig:Dlayers}
\end{figure}
\\
\textbf{Confusion matrices: } Figure~\ref{fig:CMs} shows the confusion matrices of the validation set of RAF-DB for ResNet50, ViT and ViT+SE. ViT and ViT+SE have better performance on all expressions except the Happy expression compared to ResNet50 performance. Although, the ViT+SE is 0.19\% more accurate than ViT, it only outperforms in 4 facial expressions out of 7 basic expressions, which are Fear, Happy, Sad and Surprise. The ViT performs better in Angry, Disgust and Neutral expressions. 

\newpage
\section{State-of-the-art}
\textbf{Survey on the used databases:} Table~\ref{tab:databases} shows an overview of the facial experiments databases that are used in our experiments.
\\
\textbf{Summary of state-of-the-art:} In Table~\ref{tab:SOTA} we summarize different proposed solutions in literature into 3 different approaches: Handcrafted, Hybrid and Deep Learning. The Table gives details about the year, the core of the proposed architecture and the databases used for the evaluation.
\\
\begin{table}[!ht]
  \caption{Survey on databases of Macro-Expressions. BE: Basic Expressions, CE: Compound Expressions, Publ.: Publications, Condit.: Conditions.} 
  \label{tab:databases}
  \centering
  \begin{tabular}{r|m{1.1cm}m{0.7cm}|m{2.3cm}m{1cm}m{1.7cm}m{1cm}}
    \toprule
    Database  & Publ. & Year&Annotation & Condit. & Data format &  Classes   \\
    \midrule
    \midrule
    CK+~\cite{5543262}  & CVPRW & 2010 & 593 sequences$^{*\dagger}$ & Lab &  $640\times490$, $640\times480$ & 8BE$^{\ddagger}$  \\
    \midrule
    JAFFE~\cite{670949} &FG& 1998 & 213 images$^{\dagger}$ & Lab &  $256\times256$ & 7 BE \\
    \midrule
    FER-2013~\cite{carrier2013fer} & ICONIP& 2013 & 35,887 images$^{\dagger}$ & Web & $48\times48$ & 7 BE \\
    \midrule
    SFEW~\cite{6130508} & ICCV& 2011 & 1,766 images$^{*}$ & Movie &  $720\times576$ & 7 BE \\
    \midrule
    RAF-DB~\cite{li2017reliable} & CVPR& 2017 & 29,672 images$^{*\dagger}$ & Internet &  $256\times256$ & 7BE \\
    
    \bottomrule
    
  \end{tabular}
  \\
  $^{\dagger}$ Gray scale, 
    $^{*}$ RGB,
    $^{\ddagger}$ 7BE + Contempt
\end{table}
\begin{table}[!ht]
  \caption{Summary of representative approaches for facial expressions recognition.}
  \label{tab:SOTA}
  \centering
  \begin{tabular}{lr|ll|m{2cm}m{4cm}}
    \toprule
    & Methods & Publ. & Year &Architecture&   Databases    \\
    \midrule
    \midrule
    \parbox[t]{2mm}{\multirow{6}{*}{\rotatebox[origin=c]{90}{Handcrafted}}} & \cite{506414} & TPAMI& 1996& OF& Private database \\ 
    \cmidrule(r){2-6}
      &\cite{7026204} & ICIP& 2014& PHOG, LPQ&  CK+\cite{5543262} \\
      \cmidrule(r){2-6}
     &\cite{6998925} & Trans. AC. & 2015 & LBP&  JAFFE\cite{670949}, CK+\cite{5543262}  \\
     \cmidrule(r){2-6}
     &\cite{9378702} & IHSH & 2020 & LBP, HOG&  JAFFE\cite{670949}, KDEF\cite{kdef51}, RafD\cite{rafd52}  \\
     
    \midrule
    \midrule
     
     & \cite{Otberdout2018DeepCD} & BMVC& 2018 & CNN&  Oulu-CASIA\cite{4761697}, SFEW~\cite{6130508}\\
      \cmidrule(r){2-6}
      \parbox[t]{2mm}{\multirow{6}{*}{\rotatebox[origin=c]{90}{Deep learning}}}&\cite{Wang2020RegionAN} & Trans. IP.& 2020 & CNN &FER-2013\cite{carrier2013fer}, RAF-DB\cite{li2017reliable}, SFEW\cite{6130508}, AffectNet\cite{8013713} \\
     \cmidrule(r){2-6}
     &\cite{Farzaneh_2021_WACV} & WACV& 2021& CNN &  RAF-DB\cite{li2017reliable}, AffectNet\cite{8013713} \\
     \cmidrule(r){2-6}
     & \cite{Minaee2021DeepEmotionFE} & Sensors & 2021 & CNN &FER-2013\cite{carrier2013fer}, CK+\cite{5543262} , FERG\cite{aneja2016modeling}, and JAFFE\cite{670949} \\
     \cmidrule(r){2-6}
     & \cite{Shi2021LearningTA} & arXiv & 2021 & CNN &FER-2013\cite{carrier2013fer}, RAF-DB\cite{li2017reliable},     AffectNet\cite{8013713} \\
     \midrule
     \midrule
     & \cite{article1} & ICMI& 2015 & LBP, CNN &   EmotiW 2015\cite{SFEW2015} \\ 
     \cmidrule(r){2-6}
     \parbox[t]{2mm}{\multirow{3}{*}{\rotatebox[origin=c]{90}{Hybrid}}} & \cite{9084763} & ITNEC& 2020 & LBP, CNN &  FER-2013~\cite{carrier2013fer}  \\ 
     \cmidrule(r){2-6}
     &\cite{Ma2021RobustFE} & arXiv & 2021 & LBP, CNN, ViT &FERPlus~\cite{Barsoum2016TrainingDN},  RAF-DB~\cite{li2017reliable}, AffectNet~\cite{8013713},  CK+~\cite{5543262} \\
    \bottomrule
  \end{tabular}
\end{table}


\end{document}